\documentclass[runningheads]{llncs}
\usepackage{caption}
\usepackage{subcaption}
\usepackage{graphicx}
\begin{document}

\title{Label quality in AffectNet: results of crowd-based re-annotation}
\titlerunning{Re-annotating AffectNet}
%
\author{Doo Yon Kim\inst{1}\orcidID{0000-0003-0701-3931} \and
Christian Wallraven\inst{2,*}\orcidID{0000-0002-2604-9115} 
}
\authorrunning{Kim and Wallraven}
%
\institute{Department of Artificial Intelligence, Korea University, Seoul, Korea\inst{1} \newline
\email{kdoodoo@korea.ac.kr}
\newline
Department of Artificial Intelligence \& Department of Brain and Cognitive Engineering, Korea University, Seoul, Korea\inst{2}  \newline
\email{wallraven@korea.ac.kr}
}
\maketitle              
\begin{abstract}
AffectNet is one of the most popular resources for facial expression recognition (FER) on relatively unconstrained in-the-wild images. Given that images were annotated by only one annotator with limited consistency checks on the data, however, label quality and consistency may be limited. Here, we take a similar approach to a study that re-labeled another, smaller dataset (FER2013) with crowd-based annotations, and report results from a re-labeling and re-annotation of a subset of difficult AffectNet faces with 13 people on both expression label, and valence and arousal ratings. Our results show that human labels overall have medium to good consistency, whereas human ratings especially for valence are in excellent agreement. Importantly, however, crowd-based labels are significantly shifting towards neutral and happy categories and crowd-based affective ratings form a consistent pattern different from the original ratings. ResNets fully trained on the original AffectNet dataset do not predict human voting patterns, but when weakly-trained do so much better, particularly for valence. Our results have important ramifications for label quality in affective computing.  

\end{abstract}

\section{Introduction}
Despite recent advances in facial expression recognition (FER) from images, FER on so-called in-the-wild images taken in less constrained contexts still has much lower performance when compared with FER on controlled datasets. Recent performance on one of the most popular in-the-wild datasets - AffectNet \cite{Mollahosseini2019} - is "only" around 61\% \cite{Li2020}. This performance falls far short of performance levels on controlled datasets, such as the CK+ dataset \cite{ck} with 99.69\% \cite{ck_perform} and other in-the-wild datasets, such as the (smaller) FER+ dataset \cite{fer+} with 89.75\% \cite{fer+_perform}. For a more in-depth review of the current state-of-the-art in deep-learning-based FER, see \cite{SOTA,Multimodal,Wang2020,Albanie2018,Mostafa2021}.

Interestingly, for FER+, an earlier version that was annotated by the original dataset creators \cite{fer_original} called FER2013, had a much lower recognition baseline of 70.22\%. Whereas the FER2013 dataset was annotated by only one individual, a follow-up study \cite{fer+} re-labeled the data based on crowd annotation and showed that the resulting maximum vote label also provided a better computational recognition performance. 

Since label quality directly determines performance outcomes, providing good and consistent labels has been a recent focus in the machine learning field. Performance increases can not only come from more advanced architectures, but much more "trivially", from annotating the data with clean, correct labels. Issues of label quality have, for example, been highlighted in the recent work by \cite{mit}---the authors showed that there is an average of 3.4\% errors in several of the surveyed datasets, and that by omitting 6\% of correctly labeled images, a smaller ResNet18 can perform better than a larger ResNet50. 

\begin{figure}
\centering

\includegraphics[width=\textwidth]{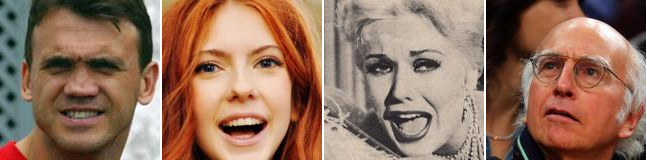}
\caption{Four images of AffectNet (originally labeled as anger, surprise, happy, and sad (from left)) illustrating the challenges in uniquely labeling expressions.} \label{fig1}
\vspace{-0.7cm}
\end{figure}

Given the relatively lower levels of performance on AffectNet, here we wanted to revisit this dataset with a similar approach to that taken to create FER+. As an example of the potential issues with AffectNet, see Figure \ref{fig1}, which shows four examples of images from AffectNet that seem to have problematic labels. Given the large size of AffectNet, we here first report results of a pilot test that uses a subset of (difficult-to-recognize) images of the different expression categories of AffectNet. We re-labeled these both in terms of expression label and also in terms of affective rating of valence and arousal by 13 naive annotators. We compare the consistency of annotators and also how well the images fit to the predictions made by a deep-learning model naively trained on the original AffectNet before and after crowd-relabeling.

\section{Experiment}
Originally, AffectNet is composed of 11 different categories that also include labels such as "No face", for example. Here, we solely focused on the seven emotional expressions plus the neutral category, resulting in: Neutral, Happy, Sad, Disgust, Surprise, Contempt, Fear, and Anger as labels. From each expression, 100 potentially-confusing images were selected by pre-screening the original images, resulting in a total of 800 images for re-annotation. 

\subsection{ResNet50}
As our deep neural network (DNN) backbone, we used a standard, ImageNet pre-trained ResNet-50, which is one of the most widely used standard architectures in image classification tasks \cite{he2016}. The architecture is based on 50 blocks of convolutional filters of size 3 X 3 with residual connections that allow the optimizer to skip a whole convolutional layer, thereby increasing the effectiveness of the total network. 

Implementation details for training the ResNet50 architecture on AffectNet were as follows: for augmentation, we added intensity normalization of the color data into 0 to 1, as well as clockwise and anti-clockwise rotations of up to 10 degrees, affine shearing up to 0.1, as well as horizontal flips. Class weights (given the imbalance in label categories) were used for a weighted cross-entropy loss that was optimized with Adam at an initial learning rate of 0.0001.  

This model was trained with 256x256px color images of the AffectNet training set of 287,561 images for 50 epochs and a batch size of 64, after which it reached a validation accuracy of 53.15\% (ResNet50 50 epoch) - its best validation accuracy throughout the run, however, was 59.25\% (ResNet50 Best), which is only less than two percent worse than the state-of-the-art in AffectNet on eight classes with a much more involved architecture\cite{Li2020}. To look at the early performance of this model, we also saved its snapshot after the first epoch (ResNet50 1 epoch).

Finally, we set up another training scheme that would perhaps be more akin to human learning, in which one epoch used only 512 randomly-chosen images, but we trained for much longer (1000 epochs). Interestingly, this very weakly-trained model achieved also a reasonable accuracy of around 53.5\%. All models are compared in  Table \ref{tab:resnet_4}. 

For valence and arousal, we followed the same model structure and created two regression models for each rating based on mean squared error (MSE) loss. These models were trained for 10 epochs (again, relatively weakly-trained), and resulted in comparable validation losses of 0.0177 (valence) and 0.0167 (arousal).
\vspace{-0.7cm}

\begin{table}
\caption{Parameters and validation performance for our ResNet50 architectures.}\label{tab:resnet_4}
\footnotesize
\begin{tabular*}{1.0\linewidth}{@{\extracolsep{\fill}}lcccc}
\hline
Model            &  Epoch  & Pre-trained  & Steps Per Epoch &Val Accuracy/Loss \\
\hline
ResNet50 50Epoch &   50    &  ImageNet    &  287,651 / batch size &     0.5315\\
ResNet50 Best    &   50    &  ImageNet    &  287,651 / batch size &  0.5925 \\
ResNet50 1Epoch  &    1    &  ImageNet    &  287,651 / batch size  &   0.4972 \\
ResNet50 8Step   &  1000   &  ImageNet    &   8    &  0.5353 \\
\hline
ResNet50 Valence   &  10   &  ImageNet    &    287,651 / batch size    &  0.0177 \\
ResNet50 Arousal   &  10   &  ImageNet    &    287,651 / batch size    &  0.0167 \\
\hline
\end{tabular*}
\vspace{-0.7cm}
\end{table}



\subsection{Selecting 800 images for re-annotation}
In total, there are 420,299 images in AffectNet. For our pilot exploration of label quality, we first pre-screened 100 images from each class to have a more manageable dataset size. To pick these images, we designed a HTML-based GUI that allowed us to quickly loop through all images for a class and select those exemplars that the trained annotator deemed "confusing".

 
\begin{figure}
\includegraphics[width=\textwidth]{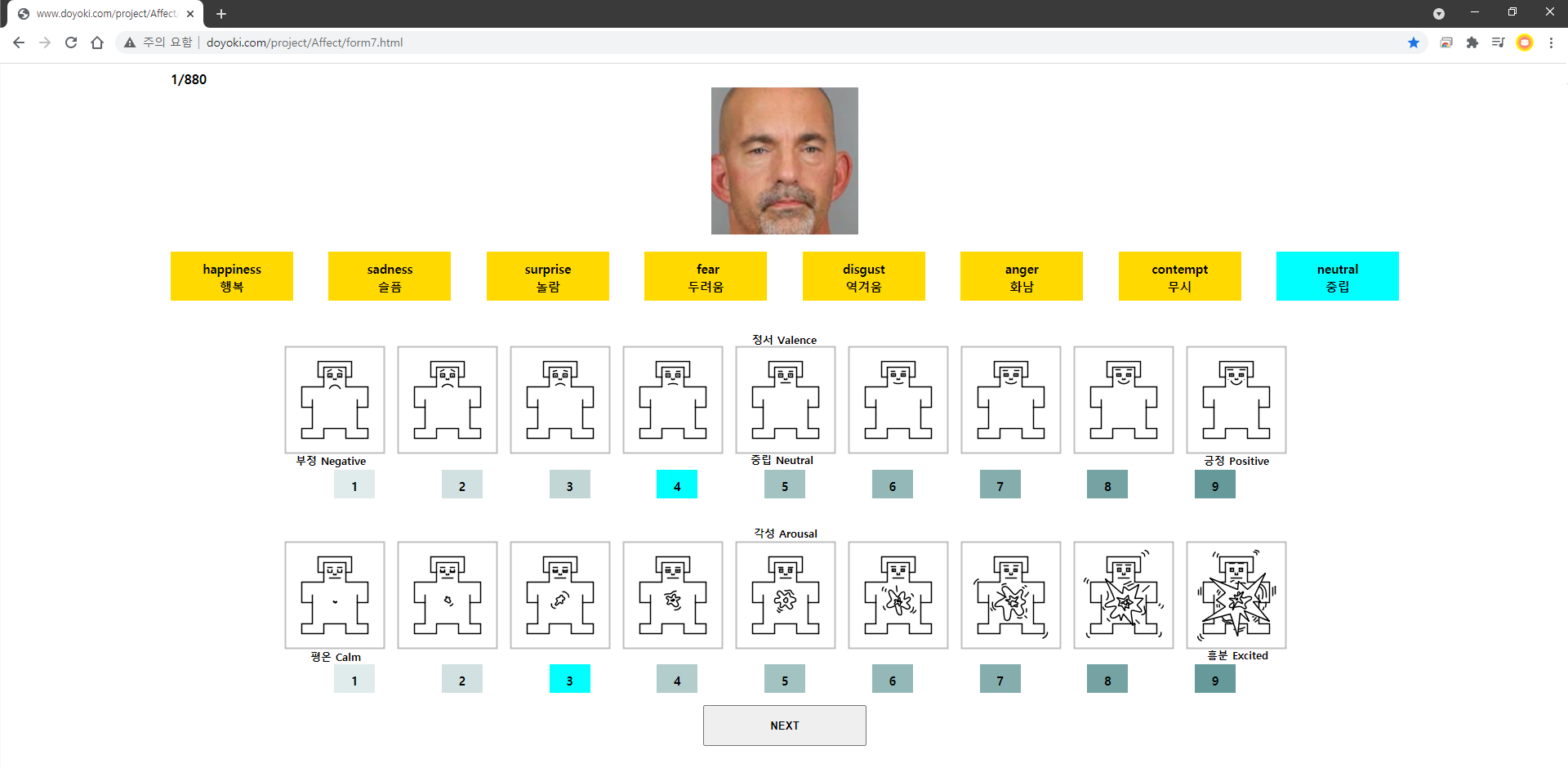}
\caption{Screenshot of the annotation GUI for the main experiment showing the image, the category selector and the SAM manikins for affective ratings.}
\vspace{-0.7cm}\label{fig:expscreenshot}
\end{figure}

\subsection{Annotation experiment}
For the experiment, we designed a software application based on HTML shown in Figure \ref{fig:expscreenshot}. For each image, participants were asked to make three annotations: in the first row, they were to click on one of the eight categories that best described the expression content of the image; the second and third row showed self-assessment manikins (SAM, \cite{sam}) for rating of valence and arousal at 9 different rating levels. 

The experiment was conducted in a quiet room and supervised in real-time by the experimenter. Participants were first instructed about the experimental procedure and were familiarized with expression labels and valence and arousal ratings. The experimenter also demoed the experiment with one dummy picture to instruct participants about how to pick the expression, to rate valence and arousal, and to move on to the next picture. Unlike AffectNet, we did \emph{not} constrain valence and arousal values to pre-defined ranges (see \cite{Mollahosseini2019}), so as not to affect participants' intuitive evaluations of the affective content.

We recruited a total of N=13 people. (7 female, 6 male, mean age (STD) = 33.23 (12.39) years) from the population of Korea University. Participants were naive as to the purpose of the experiment, and reported no neurological or psychological issues that would interfere with emotion processing.

\section{Re-annotation results}
\subsection{Expression categories}
The total number of votes added up to 10,400---as a first, omnibus result, we observed that of these votes, 8,658 (83.25\%) did \emph{not} agree with the original AffectNet labels, indicating that these labels may have issues.

\begin{figure}
    \begin{subfigure}[t]{0.46\textwidth}
      \includegraphics[width=\textwidth]{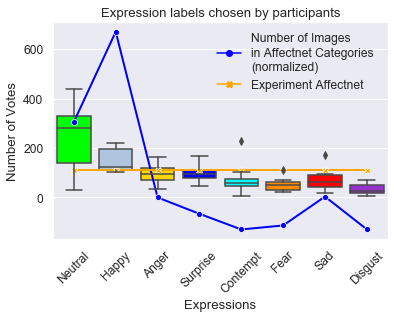}
      \caption{Box-whisker plot showing the results of the human re-annotation in comparison to the constant AffectNet label distribution (orange line) and the distribution of all images in AffectNet (blue line).}\label{fig:expplot}
    \end{subfigure}
    \hfill
    \begin{subfigure}[t]{0.46\textwidth}
      \includegraphics[width=\textwidth]{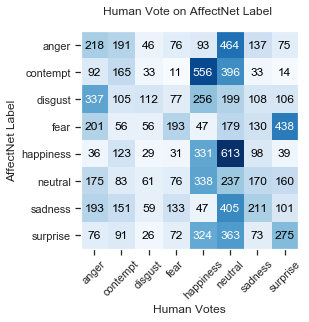}
      \caption{Confusion-type matrix showing which label participants chose depending on the original AffectNet label.}\label{fig:avgmax}
    \end{subfigure}
    \caption{Voting results and confusion matrix}
  \end{figure}

Figure \ref{fig:expplot} shows the overall results for participants' votes\footnote{In determining the final vote to be counted, we also experimented with different kinds of maximum (modal) voting, such as maximum vote across all votes, maximum votes for each image with different ways of breaking ties (counting the first tie, the second, or the third). We found that the average voting pattern was not affected by these methods, and hence used this pattern in the remainder of this paper.} in more detail (as box plots). The original AffectNet data was uniformly distributed across the eight classes (Figure \ref{fig:expplot} orange line)---this is clearly not the case for the median votes, however: On average, participants chose a much higher proportion of neutral and happy labels and much lower proportions of contempt, fear, sad, and disgust. We also observed variability across expression categories in the voting patterns, especially for the neutral, and to lesser degrees for the happy and sad categories, indicating that there was less agreement among participants for these category labels.

The overall consistency of participants on voting was evaluated by taking each participant's individual voting pattern per expression and correlating this with all other participants' voting patterns. The average value of the upper triangular part of the resulting correlation matrix is a measure of \emph{relative} consistency and was determined to be $r_{expr}=.550$, indicating medium to good consistency levels. 



Figure \ref{fig:expplot} also shows the original number of images in each category in AffectNet (blue line, normalized). We can see that this distribution---as an extremely weak measure of a "real-life" distribution of these facial expression categories---fits somewhat better to the median voting pattern (the correlation value for the human voting data to this data is $r=0.71$, see also Figure \ref{fig:humanmachine_corr} below). This may be an indication that participants implicitly fitted their overall voting strategy to some sort of "hidden distribution"; for example, one would expect that most of the time, faces display no strong facial emotion and hence will be neutral. The next most common expression may be happy followed by the other categories with contempt and disgust being perhaps the least commonly-encountered expressions.

Figure \ref{fig:avgmax} displays a type of confusion matrix, which shows how the original AffectNet labels and the maximum vote labels are connected. As can be seen, the matrix does not exhibit any diagonal structure owing to the large number of differing votes we received. Many contempt images, for example, got relabelled as happy or neutral. At the same time, and in line with the results in Figure \ref{fig:expplot}, many images originally annotated with an emotional label received votes for the neutral expression. 

Overall, our results so far on the 800, pre-screened images suggest that the original expression labels in AffectNet for these difficult images are far from ideal. 

\subsection{Valence and Arousal}
\begin{figure}
\centering
a)\includegraphics[width=0.47\textwidth]{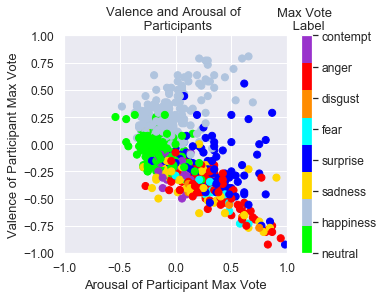}
b)\includegraphics[width=0.47\textwidth]{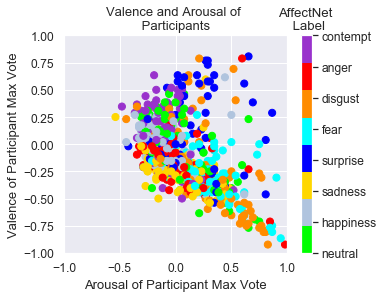}
\caption{a) Arousal-Valence space colored with our max vote labels. b) Same space colored with original AffectNet labels} \label{fig:valar}
\end{figure}

To better compare our data with those from AffectNet, we mapped the 1-9 scale into an -1.0 to 1.0 range. Figure \ref{fig:valar}a) shows the full data of our experiment with our arousal and valence ratings colored by the maximum vote labels. First of all, we see a clear "rotated U"-shape that is reminiscent of the data from affective ratings obtained for the International Affective Picture System (IAPS, \cite{IAPS})---in other words, as valence increases positively and negatively arousal increases into the positive direction in both cases. This pattern is a well-established result for affective ratings of pictures and videos \cite{IAPS,castillo2014semantic}. 

In this context, it is important to mention that the AffectNet annotations for valence and arousal were collected in a different way compared to ours: their annotation scheme used a software application in which the annotator clicked into a pre-defined region within a two-dimensional coordinate system (see \cite{Mollahosseini2019}). In our case, the two annotations were obtained in a more "traditional" fashion using the SAM method and without restricting participants to certain value ranges, depending on the label, which may explain some of the differences in rating pattern.

One of the reasons for the different annotation scheme for the original AffectNet ratings may have been to ensure a better consistency for valence and arousal evaluations. To look more closely into this matter, Figure \ref{fig:valar} also shows coloring based on the expression categories obtained from the maximum votes, which clearly fit this valence/arousal space well: the green neutral expression stays near the center, whereas red anger and grey happiness clearly move to  the upper positive and bottom negative quadrants of the space. If we change labels to the original AffectNet labels, however, the space becomes much more scattered and less consistent as shown in Figure \ref{fig:valar}b. Hence, the overall pattern of ratings seems consistent with the voted labels in our experiment.

\begin{figure}
\centering
a)\includegraphics[width=0.47\textwidth]{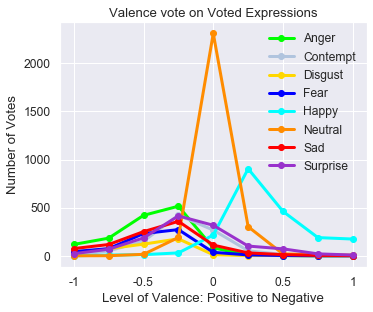}
b)\includegraphics[width=0.47\textwidth]{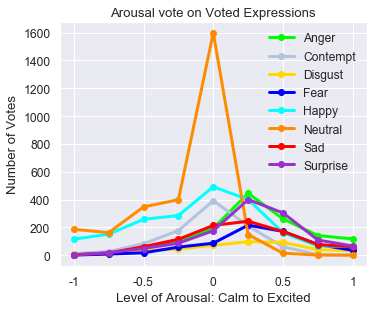}
\caption{Human votes on a)valence and b)arousal for voted expressions. In Figure \ref{fig:aro_val_vote_to_votes}, it shows how people annotated valence and arousal related to their votes to the expressions. For example, we can clearly see when people voted for 'neutral' they also voted mostly on middle point of both valence and arousal. } \label{fig:aro_val_vote_to_votes}
\end{figure}

To reinforce this point, in Figure \ref{fig:aro_val_vote_to_votes} we show the overall distribution of valence and arousal votes when labeled with the chosen, maximum vote. Here, the distributions also have peaks at expected values (neutral being at 0, happy peaking at positive valence values, and anger peaking at positive arousal values). 

\begin{figure}
\centering
a)\includegraphics[width=0.47\textwidth]{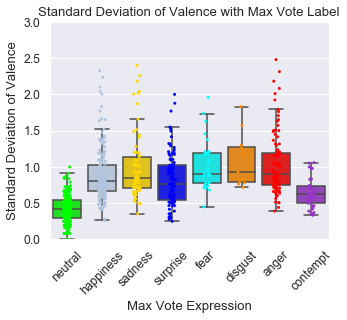}
b)\includegraphics[width=0.47\textwidth]{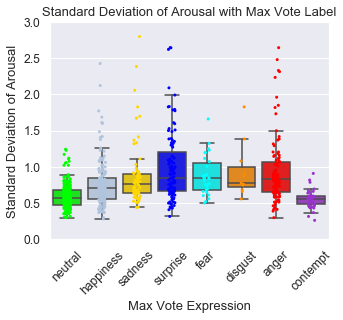}
\caption{Standard deviations of a) valence and b) arousal ratings across participants based on the chosen maximum vote label.} \label{fig:std_aroval}
\end{figure}

We next analyzed the standard deviation of valence and arousal ratings within each category selected by participants---these results are shown in Figure \ref{fig:std_aroval}a,b. As can be seen, overall standard deviations were in the range from 0.2 to 1.0 and were lowest for neutral and contempt for both valence and arousal. Valence seemed to have similar variability for each expression compared to arousal ratings. Overall, \emph{absolute} agreement on the rating values was relatively high at these low levels of variability.

\begin{figure}
\centering
a)\includegraphics[width=0.47\textwidth]{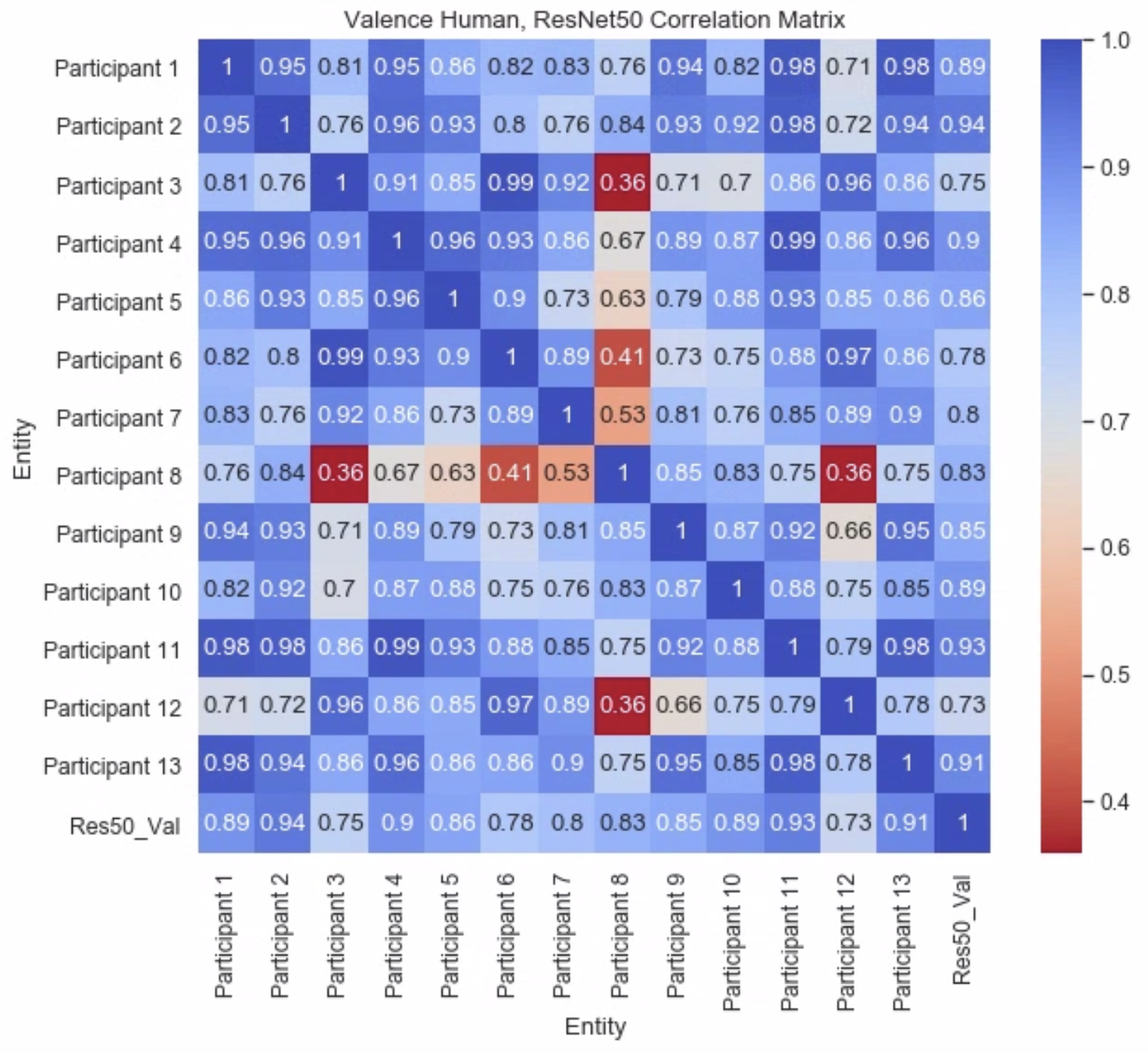}
b)\includegraphics[width=0.47\textwidth]{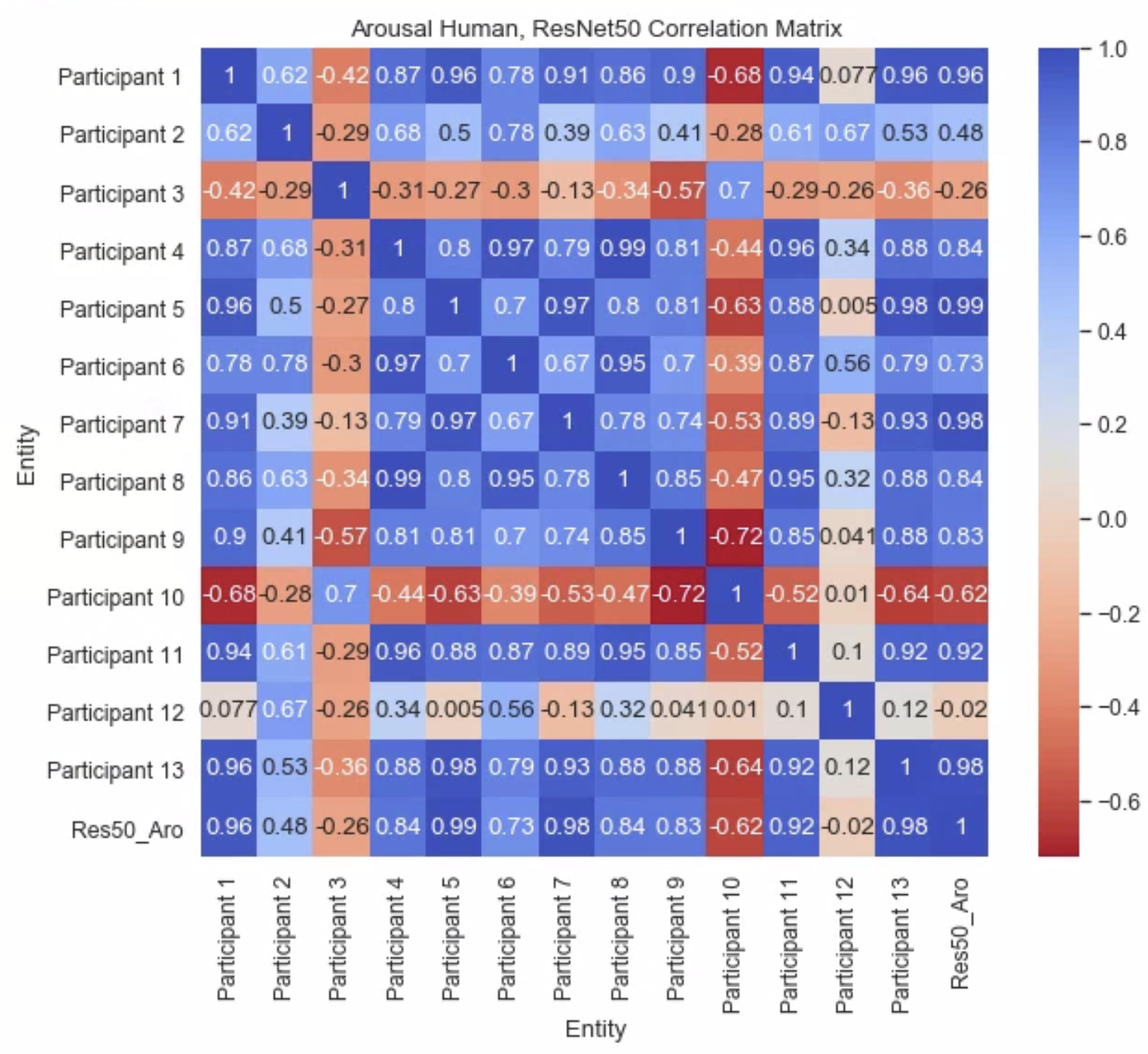}
\caption{Correlation matrix between participants for a) valence and b) arousal ratings. The matrix also includes an extra row for the ResNet model (see analysis below).} \label{fig:valar_corr}
\end{figure}

We again analyzed the overall correlations among participants in the same manner as done for the expressions. These matrices are shown in Figure \ref{fig:valar_corr}. The average correlations measuring consistency in \emph{relative} voting pattern among human participants was $r_{val}=.865$ for valence and $r_{aro}=.468$ for arousal. Relative agreement was very high for valence, but only at medium levels for arousal due to some participants choosing a different rating profile (see Figure \ref{fig:valar_corr}b, Participant 10, for example). Our results show that participants can agree much better on valence ratings of the expressions than on the label---arousal is at similar levels of relative consistency as the labels.


\section{Humans vs. DNNs}
\subsection{Expressions}
\begin{figure}
    \begin{subfigure}[t]{0.46\textwidth}
      \includegraphics[width=\textwidth]{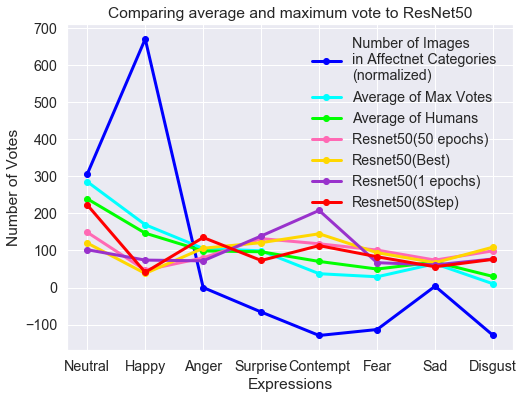}
      \caption{Average voting for the four different ResNet models compared to the average human vote (green) and the number of images in AffectNet categories (blue).}\label{fig:compare_med}
    \end{subfigure}
    \hfill
    \begin{subfigure}[t]{0.46\textwidth}
      \includegraphics[width=\textwidth]{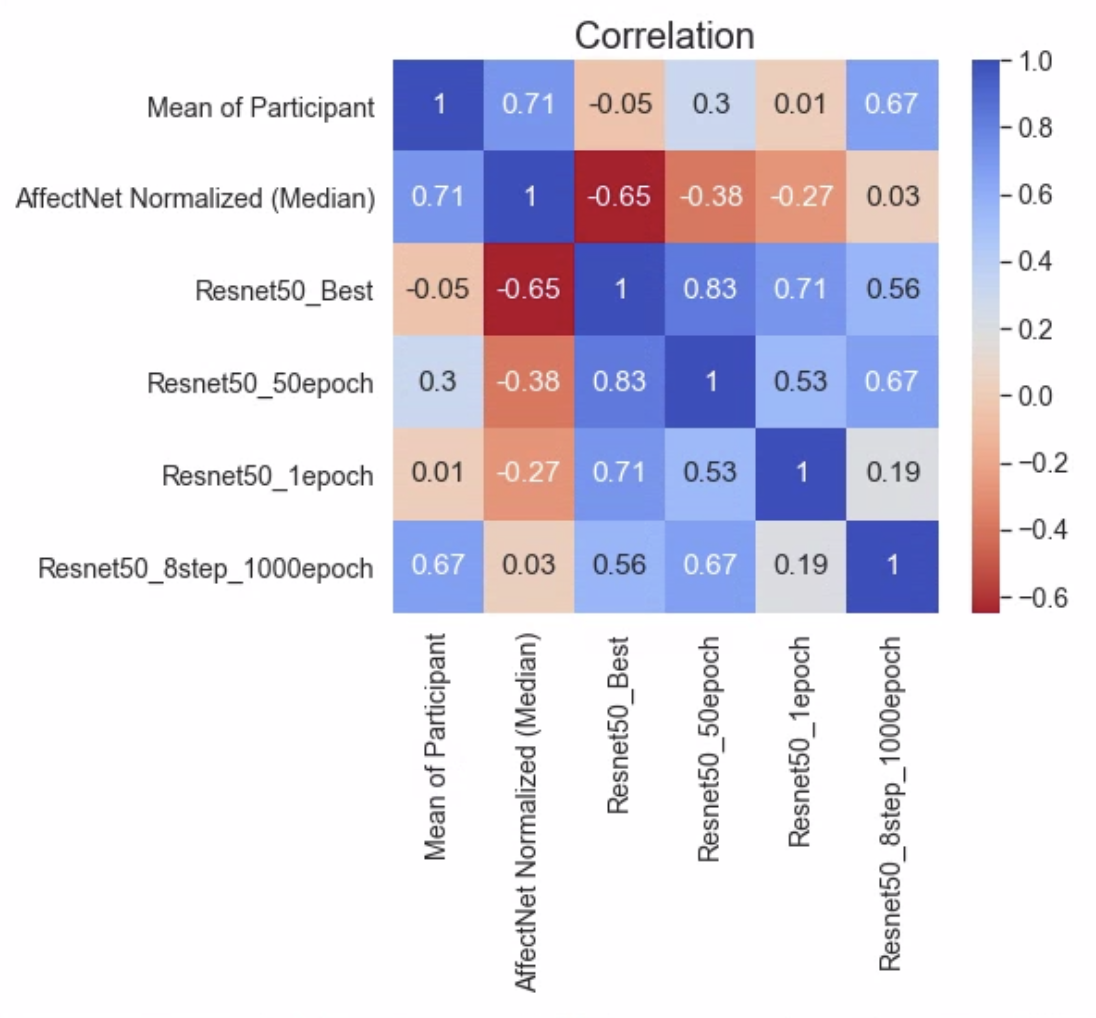}
      \caption{Correlation matrix showing similarities in voting pattern for all curves in Figure \ref{fig:compare_med}}\label{fig:humanmachine_corr}
    \end{subfigure}
    \caption{Comparison of human and machine voting patterns.}
  \end{figure}

 Figure \ref{fig:HDNN} compares voting patterns of the different DNNs we trained to the human re-annotation results. Figure \ref{fig:compare_med} shows the different voting patterns across categories (line-plots were chosen to highlight the overall pattern similarity). At first glance, none of the ResNets' predictions matches the human pattern perfectly---however, when looking at Figure \ref{fig:humanmachine_corr}, which plots the correlation values among all voting patterns, we can see that the most similar voting pattern to human performance is obtained for the ResNet trained very weakly with only 8 images for 1000 epochs. This is at similar levels to the correlation for human votes to the AffectNet label distribution (blue line).
 
 Overall, we find that ResNets trained on the AffectNet original label distribution cannot capture the human voting pattern well, except for a model that was only weakly-trained (yet still achieved a relatively high performance on the AffectNet validation set).


\label{fig:HDNN}

\subsection{Valence and Arousal }
\begin{figure}
\centering
\begin{subfigure}[b]{0.49\textwidth}
\centering
\includegraphics[width=\textwidth]{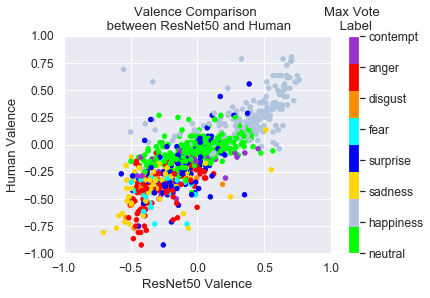}
\caption{Human max vote label.} \label{fig:val_res_hum__humlab}
\end{subfigure}
\begin{subfigure}[b]{0.49\textwidth}
\includegraphics[width=\textwidth]{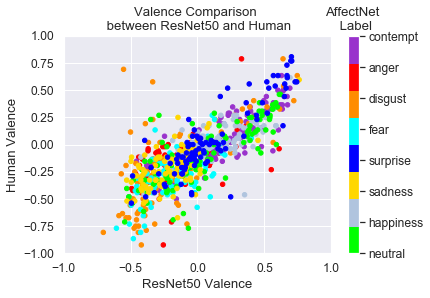}
\caption{AffectNet label} \label{fig:val_res_hum_afflab}
\end{subfigure}
\caption{Valence Comparison between ResNet50 and humans.}
\vspace{-0.7cm}
\end{figure}

We found that agreement with human valence ratings was very high for the simple ResNet we trained for only 10 epochs as shown in the scatter plot in Figure \ref{fig:val_res_hum__humlab}. The coloring in this figure was done according to the human maximum voting label and shows the expected distribution of expression categories along the valence range (from angry in red to happy in grey). Conversely, if we color the same data according to the original AffectNet labels, the plot looks much less well-structured (Figure \ref{fig:val_res_hum_afflab}).

\begin{figure}
\centering
\begin{subfigure}[b]{0.49\textwidth}
\centering
\includegraphics[width=\textwidth]{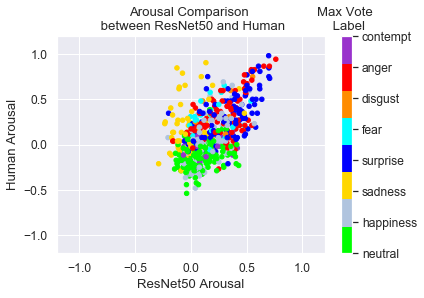}
\caption{Human max vote label.} \label{fig:aro_res_hum__humlab}
\end{subfigure}
\begin{subfigure}[b]{0.49\textwidth}
\includegraphics[width=\textwidth]{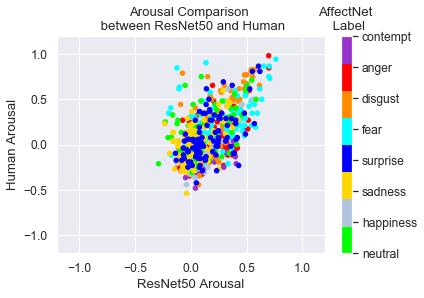}
\caption{AffectNet label} \label{fig:aro_res_hum_afflab}
\end{subfigure}
\caption{Arousal Comparison between ResNet50 and humans.}
\vspace{-0.7cm}
\end{figure}

Results for arousal also show good agreement of the ResNet model with human data (Figure \ref{fig:aro_res_hum__humlab}). Again, the human maximum vote coloring produces the desired relationship of arousal to expression category label (cf. neutral at the bottom part to anger and surprised expressions towards the top). Compared to the valence data, however, arousal seems to have more spread---this may be due in part to the fact that arousal values did not span the whole scale, which could have resulted in worse generalization performance for the network. Again, Figure \ref{fig:aro_res_hum_afflab}, which uses the original AffectNet labels does not produce consistent relationships between ratings and expression categories.

The last rows in Figure \ref{fig:valar_corr} show the correlations of the trained ResNet models with the human ratings. The overall, average correlation of the DNN with valence is $r_{val,DNN}=.866$ and for arousal $r_{aro,DNN}=.504$. These results confirm the previous analysis and show that, indeed, the ResNet with minimal training of 10 epochs is capable of capturing the valence ratings on our newly-annotated images well. The relative agreement with the arousal ratings is at medium levels---similar to human-human consistency values obtained above.

\section{Discussion and conclusion}
Given the recent discussion in the field about label quality in datasets \cite{mit} and the "success" that crowd-based re-annotation had for the FER+ dataset, here we presented results of re-annotation experiments on the popular AffectNet dataset. 

Thirteen participants re-annotated our pre-screened dataset of 800, difficult-to-parse images with expression category and affective ratings. Overall, we found that expression category votes agreed only very weakly with the original AffectNet labels: only 13\% of votes kept the same label. Indeed, participants instead seemed to follow an implicit bias to real-life frequencies in their voting with large numbers of votes falling into the neutral category, for example. Adding the neutral category is relatively rare in human annotation experiments as participants usually are forced to choose one out of seven emotional categories from among expressive faces - this may bias the results, however, as shown, for example, in \cite{Russel1992,Nusseck2008} as participants cannot select a "non-signal". It will be interesting, nonetheless, to repeat the annotation with different amounts of categories or with a ranking strategy (similar to the TopN-accuracy results reported for large-scale classification tasks). Overall, our results so far clearly confirm the difficulty to assign unique labels to facial expressions out-of-context (that is, without temporal or auditory contextual information)---an issue that was also visible in the large variability in some of the expression categories. Importantly, this issue was also highlighted in a recent review paper in affective psychology \cite{FeldmanBarrett2019}, which cast doubt on context-free, unambiguous labeling of facial expressions. 

The issue of labels is circumvented to some degree by the affective ratings we obtained in the annotation. Here, participants had less variability overall, but also showed clear, systematic, and consistent deviations from the original AffectNet ratings. In line with a large number of other results \cite{IAPS,castillo2014semantic}, we obtained a valence-arousal space that showcased dependencies between the two dimensions. This may in part be due to the difference in methods, where the AffectNet annotations placed implicit and explicit restrictions on participants' rating pattern---a topic that will be interesting to follow up in future studies. The observed rating patterns, however, did match well with the average expression categories and were consistent among participants with little variability.

Finally, we compared the human crowd annotations with predictions of DNNs. In terms of expression categories, we found no good match, except for a weakly-trained, somewhat "unconventional" DNN that showed good correlations. Given that the human data did not match well with the AffectNet labels, the failure of the AffectNet-trained DNNs to capture human voting patterns is perhaps not surprising. Only a much more extensive re-annotation of AffectNet will be able to shed final light on potential performance gains (and better fitting quality), as here we were only able to test a first set of 800 images. This work is currently underway.

In terms of valence and arousal, however, we found much better agreement between the ResNets and human performance, especially for valence but also to some degree for arousal. Since the number of networks we tested for these ratings, was limited, however, we can only cautiously suggest that this could be due to the fact that affective ratings for our (difficult) images suffered less from ambiguity compared to the labels (see human results). Future work will need to test the degree to which the weakly-trained networks used in our work are better at capturing the label- and/or rating space of human annotations. 

Overall, our results clearly highlight issues with expression labeling and point to the need for the field to use also other, continuous annotation schemes (like valence and arousal, but more evaluative dimensions are possible \cite{dilara2020}) and/or focus on analysis of affective data in more contextually rich environments.

\section{Acknowledgments}
This work was supported by Institute of Information Communications Technology Planning Evaluation (IITP; No. 2019-0-00079, Department of Artificial Intelligence, Korea University) and National Research Foundation of Korea (NRF; NRF-2017M3C7A1041824) grant funded by the Korean government (MSIT).

\end{document}